\documentclass[sn-mathphys, iicol]{sn-jnl}

\usepackage{amsmath,amssymb,amsfonts,bm}

\jyear{2021}%

\theoremstyle{thmstyleone}%
\theoremstyle{thmstyletwo}%

\theoremstyle{thmstylethree}%

\begin{document}

\title[Article Title]{Interactive Physically-Based Simulation of Roadheader Robot}


\author[1]{\fnm{Shengzhe} \sur{Hou}}\email{housz@sdust.edu.cn}

\author*[1]{\fnm{Xinming} \sur{Lu}}\email{luxinming@sdust.edu.cn}

\author[1]{\fnm{Wenli} \sur{Gao}}\email{gaowl91@163.com}

\author[2]{\fnm{Shuai} \sur{Jiang}}\email{m211111@hiroshima-u.ac.jp}

\author[1]{\fnm{Xingli} \sur{Zhang}}\email{xlzhang\_only@163.com}

\affil*[1]{\orgdiv{College of Computer Science and Engineering}, \orgname{Shandong University of Science and Technology}, \orgaddress{\street{579 Qianwangang Road}, \city{Qingdao}, \postcode{266590}, \state{Shandong Province}, \country{China}}}

\affil[2]{\orgdiv{Graduate School of Advanced Science and Engineering}, \orgname{Hiroshima University}, \orgaddress{\street{1-3-2 Kagamiyama}, \city{Higashi-Hiroshima City}, \postcode{739-8527}, \state{Hiroshima}, \country{Japan}}}


\abstract{
Roadheader is an engineering robot widely used in underground engineering and mining industry. 
Interactive dynamics simulation of roadheader is a fundamental problem in unmanned excavation and virtual reality training. 
However, current research is only based on traditional animation techniques or commercial game engines.
There are few studies that apply real-time physical simulation of computer graphics to the field of roadheader robot.
This paper aims to present an interactive physically-based simulation system of roadheader robot. 
To this end, an improved multibody simulation method based on generalized coordinates is proposed.
First, our simulation method describes robot dynamics based on generalized coordinates.
Compared to state-of-the-art methods, our method is more stable and accurate.
Numerical simulation results showed that our method has significantly less error than the game engine in the same number of iterations. 
Second, we adopt the symplectic Euler integrator instead of the conventional fourth-order Runge-Kutta (RK4) method for dynamics iteration.
Compared with other integrators, our method is more stable in energy drift during long-term simulation.
The test results showed that our system achieved real-time interaction performance of 60 frames per second (fps).
Furthermore, we propose a model format for geometric and robotics modeling of roadheaders to implement the system.
Our interactive simulation system of roadheader meets the requirements of interactivity, accuracy and stability.
}




\maketitle

\section{Introduction}
\label{sec:intro}

Roadheader is a kind of tunneling robot, one of the most important machinery in the mining industry and underground engineering \cite{deshmukh2020roadheader}. 
Generally, underground engineering is very dangerous for people. 
To improve safety, unmanned underground tunneling \cite{li2018intelligent} and Virtual Reality (VR) based worker training \cite{grabowski2015virtual} have become a trend in recent years. 
The development of interactive computer graphics technology, especially physically-based 
robotics simulation, provides theoretical support for achieving these targets.
The dynamics-based three-dimensional (3D) visualization system can not only simulate the working state of the roadheader robot most physically but also provide a realistic user experience.
Therefore, physically-based robot simulation has become one of the core technologies of digital twin \cite{bilberg2019digital}.
However, most of the current solutions are based on kinematics animation \cite{grabowski2015virtual} or directly using game-oriented commercial engines \cite{choi118use}.
This paper presents a dynamics-based interactive roadheader simulation system that is accurate and stable.

Graphical robot simulation belongs to the interdisciplinary field of computer graphics and robotics \cite{liu2021role}. 
Interactive robot simulation aims to compute and show the robot's motion state in real-time based on articulated rigid body dynamics (usually called multibody dynamics in applied mathematics and mechanics) \cite{bender2014interactive}.
Since the 1980s, rigid body simulation has been an important topic in computer graphics. Researchers from computer science, robotics, and mechanics have made this field flourish, and then the results have been applied to various industries \cite{luckcuck2019formal, inproceedings}.
Although single rigid body dynamics are well understood in computer graphics, articulated rigid body simulation is still a very challenging research field. 
Unlike keyframe animation, physical simulation must conform to the laws of physics.
To achieve interactivity, the computing speed of the simulator must be fast enough.
Therefore, the realization of physically-based roadheader simulation has two main challenges: accuracy and real-time. 
The metrics for the former generally include position error, energy drift, etc.
The latter usually requires robot simulators to reach 60 frames per second (fps) or higher \cite{bender2014interactive}. 
The difference in preference for these two factors has led to different research fields.
In video games or other entertainment products, accuracy is sacrificed in pursuit of real-time performance. 
In contrast, mechanical engineering simulation systems usually take a long time to compute for accuracy.
The balance of accuracy and performance is an eternal topic in the field of interactive simulation.
Our goal is to achieve a sufficiently accurate robot simulation system under the premise of satisfying interactivity.

A typical robot, like the roadheader, can be modeled as an articulated rigid body system, which is composed of rigid bodies, called links, connected by joints \cite{lynch2017modern}. 
Although Newton's second law laid the foundation of dynamics, modern simulators generally do not directly use it to describe the effect of force on bodies.
As the complexity of the system increases, equations based on Newtonian mechanics will become difficult to handle.
Inspired by analytical mechanics (including Lagrangian and Hamiltonian mechanics), computer simulators generally adopt constraint-based dynamics. Briefly, the task of the simulator is to make the system meet the established constraints at each time step iteration \cite{bargteil2019introduction}.

The researches on constraint-based rigid body dynamics can be divided into two categories: the maximal coordinates formulation and the generalized (also named reduced or minimal) coordinates formulation. 
The former uses the original coordinates to represent rigid body motion and adds auxiliary conditions to define constraints. 
In contrast, the generalized coordinates formulation uses a minimal set of coordinates to describe constrained dynamics. 
The maximal coordinates methods are widely used in game engines, such as Bullet \cite{coumans2013bullet}, ODE \cite{smith2005open}, etc.
However, methods of maximal coordinates are all approximations to dynamics and are difficult to meet the requirements of robotics for accuracy and stability. 
In the field of industrial research and application, methods of generalized coordinates have many advantages: not only accuracy is higher, but computation speed is also sufficient. 
The theoretical basis of generalized coordinates methods, analytical mechanics, can more essentially describe the nature of dynamics. Therefore, algorithms based on generalized coordinates are adopted by our roadheader simulation system.

\begin{figure*}[]%
    \centering
    \includegraphics[width=0.7\textwidth]{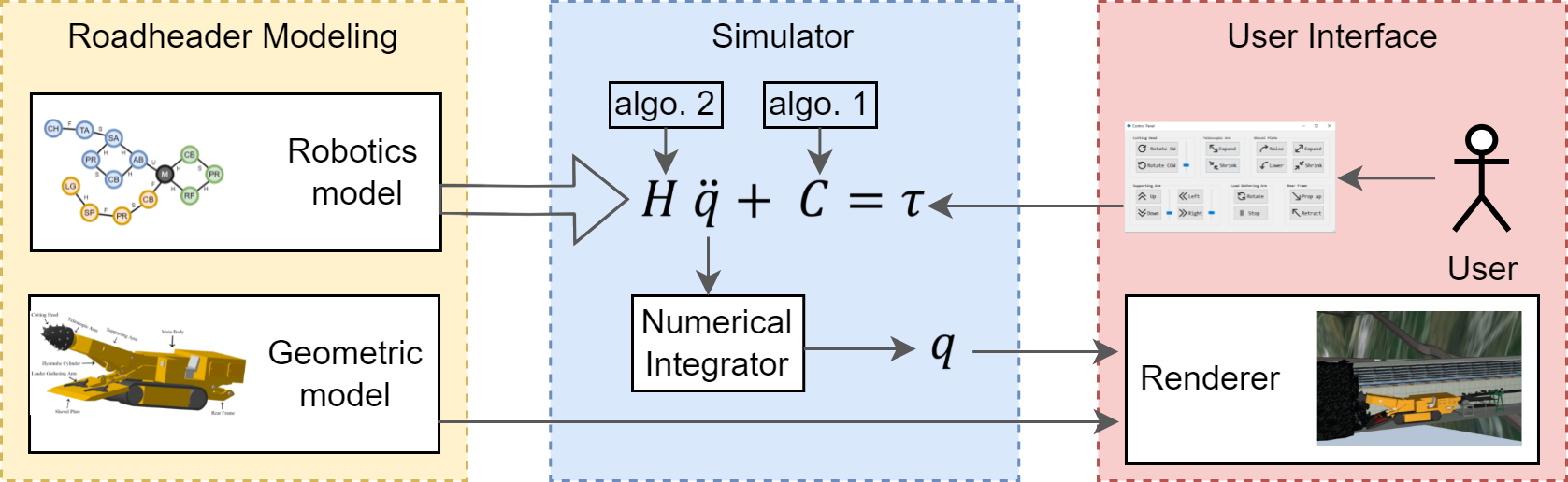}
    \caption{The framework of roadheader robot simulation system.}\label{framework}
\end{figure*}

Rigid multibody dynamics algorithms based on generalized coordinates is a key part of our roadheader simulation system.
We use space algebra based on 6-D vectors to describe related dynamic theories and algorithms. 
Traditional graphics simulation processes the linear and rotation movement of rigid bodies separately.
That will cause the simulation algorithm and implementation to be very cumbersome. 
The language of space algebra describes dynamics from a more essential perspective and is more concise.
After introducing the basic theory of space algebra, we derived the dynamic expression of a single rigid body.
Then, the equation of motion of multibody dynamics is introduced. 
The dynamic algorithms of the simulation system aim to solve the unknown variables in the equation of motion. 
Recursive Newton-Euler Algorithm and Composite Rigid Body Algorithm are adopted by our system \cite{featherstone2014rigid}. 
In recent years, the theoretical research of these algorithms is developing rapidly \cite{agarwal2014dynamics, korayem2014systematic}.
Related applications have also been studied extensively \cite{korayem2015motion}.
We apply these algorithms to real-time simulation of roadheader robot.
Numerical experiment results show that the simulation accuracy of our system is better than the solution based on a game engine (using maximal coordinates).

The type of numerical integrator is an important factor affecting the simulation stability. In each iteration of the simulation process, the integrator will integrate the current acceleration of the robot obtained by the dynamic algorithm to obtain the current position. 
An ideal simulator guarantees the conservation of energy.
However, energy drift caused by numerical error is inevitable in every integration.
In most interactive graphics applications, the fourth-order Runge-Kutta (RK4) method can make a good trade-off between performance and accuracy. Therefore, RK4 integrator has been widely used in video games and other fields. 
However, RK4 is not an energy preserving method. The energy drift generated by the RK4 method will gradually accumulate, which makes it unsuitable for long-term robot simulation.
We propose to use the symplectic Euler (also called semi-implicit Euler), an energy preserving method, as the integrator of our roadheader simulation system. 
In the energy drift test, the symplectic Euler integrator is more stable than other methods.


As shown in Figure \ref{framework}, the roadheader robot simulation system is divided into three parts: roadheader modelling, simulator, and user interface. We propose robotics model and geometric model to model physical information and graphics information of roadheader, respectively. Robotics model will be encoded by the simulator as the equation of motion for multibody system. Our goal is to get the unknown acceleration $\ddot{\boldsymbol{q}}$ . User interaction will change the right-hand term $\boldsymbol{\tau}$ . Algorithms 1 and 2 are used to solve part of this equation. The numerical integrator will get the current position ${\boldsymbol{q}}$ based on $\ddot{\boldsymbol{q}}$. The renderer will visualize the simulation results based on the current position ${\boldsymbol{q}}$ and the geometric model. 

Our contribution is mainly in three aspects: 
\begin{itemize}
  \item We propose to use the rigid multibody dynamics algorithm based on generalized coordinates to simulate the roadheader robot.
  \item A symplectic Euler integrator instead of traditional RK4 is adopted to achieve better simulation stability.
  \item A roadheader simulator is implemented to meet the requirements of interactivity, accuracy and stability.
\end{itemize}
Numerical tests show that our system meets the interactive performance requirements, and has better accuracy and stability in long-term simulations.


The paper is structured as follows.
In Section \ref{sec:relatedwork}, we briefly review the research on roadheader simulation and robot interactive algorithms.
In Section \ref{sec:methods}, we describe the theoretical basis and algorithms of rigid multibody dynamics and the principle of numerical integrators.
In Section \ref{sec:modeling}, the geometry and robotics modeling methods of the roadheader are introduced. 
In Section \ref{sec:implement}, the architecture and implementation of our roadheader simulation system are described.
In Section \ref{sec:results}, we test the system's performance, accuracy (compared with a game engine), and energy stability (compared with other numerical integrators) through numerical experiments.
Finally, conclusions of this work and future research are summarized in Section \ref{sec:conclusion}.

\section{Related Work}
\label{sec:relatedwork}

\subsection{Roadheader robot simulation}
In recent years, information technology, especially virtual reality technology, has been widely used in traditional mining engineering \cite{bellanca2019developing, van2009virtual}. 
Roadheader is an indispensable robot equipment in mining engineering. 
Improving the safety of roadheaders, especially unmanned operation and training, has been a research focus in recent years \cite{yan2019multi}. 
However, the existing research only focuses on the basic kinematics \cite{tian2018kinematic, yan2019multi} or the use of remote video to manually control the roadheader \cite{wang2017recent}, and the dynamics have not been considered.
According to the idea of digital twin, the key to efficient and safe human-robot interaction is virtual simulation of robots \cite{malik2018digital}. 
Robot simulation technology based on interactive computer graphics makes it possible to create a virtual robot environment. 

Although robot simulation technology has been widely used in video games and movies, there is little research in traditional industrial fields, especially mining engineering. 
Some studies only used simple animations without considering robot dynamics \cite{marshall2016robotics}, or simple secondary development of commercial platforms \cite{zhang2017head, andersen2020mets}. 
The system proposed in this paper is based on robot dynamics and graphical visualization techniques.
Our system is more realistic and reliable than non-physical visualization methods.
Compared to commercial physics engine based solutions, our method is more stable and accurate. 

\subsection{Dynamics modeling methods}
Robot simulation theory is derived from classic rigid body mechanics and is still active in computer graphics. 
The current robot simulation algorithms are generally divided into two categories: maximal coordinates methods and generalized coordinates methods.
In addition, some non-physical simulation methods, such as Position Based Dynamics \cite{muller2020detailed, deul2016position}, have also been applied in recent years. But the accuracy of these methods is difficult to meet the requirements of robot simulation.
Currently, the most widely used methods in the maximal coordinates can be roughly divided into two categories: penalty methods and Lagrange multipliers \cite{bender2014interactive}. Penalty methods satisfy constraints by adding spring-like penalty forces. This type of method is easy to implement but has poor accuracy. The method of Lagrange multipliers explicitly computes constraint forces to satisfy system constraints. 
The impulse-based method is similar to the Lagrangian multiplier method and is used in many game engines, such as Bullet \cite{coumans2013bullet}. 
In short, all current methods based on maximal coordinates are approximately accurate. The maximal coordinate method works well in video games, but it is difficult to meet the accuracy requirements in industrial simulators.

To achieve better accuracy and stability, our system is based on generalized coordinates algorithms instead of these maximal coordinates methods. 
The idea of generalized coordinates originated in analytical mechanics \cite{goldstein2002classical}.
Robotics and graphics have developed a variety of algorithms based on the idea of generalized coordinates \cite{featherstone2014rigid, wang2019redmax}. 
Accordingly, the mathematics of rigid body dynamics also vary \cite{murray2017mathematical}.
Our approach is rigid multibody dynamics based on spatial algebra using 6-D vectors \cite{featherstone2010beginner1}. Compared with traditional coordinate systems, our method is more concise. The dynamics algorithms are applied to the simulation of roadheader robot. In the same simulation process, our system is more accurate and stable than commercial game engine.

\subsection{Time integrators}
A system of nonlinear differential equations (ODEs) was established by the robot dynamics simulation algorithms. These differential equations are difficult to solve analytically, so numerical integrators must be used to solve them step by step.
The explicit Euler method (also known as forward Euler method) is the simplest ODE solver. Each iteration of explicit Euler integrator uses only the gradient of the system and the state of the previous step.
Although explicit Euler is highly efficient and easy to implement, it introduces additional capabilities (called numerical explosions) at each iteration \cite{hairer2006geometric}.
Implicit methods (e.g. implicit Euler methods) have higher stability, but lead to energy decay. Most importantly, implicit methods take a lot of time to compute the implicit equation, so it is not suitable for interactive simulation which requires high real-time performance \cite{house2016foundations, hairer2006geometric}.
To improve the accuracy and stability of explicit Euler method, many higher order methods have been adopted by physically-based animation field.
Currently, the fourth-order Runge-Kutta (RK4) method is widely used in virtual simulation and game engines.
While RK4 performs well in most situations, balancing performance with accuracy, it is still an explicit approach \cite{butcher2008numerical}.
That is, RK4 will inevitably introduce additional energy and lead to simulation errors.
This is not to be ignored in the long-term simulation process.
In addition, higher order methods have been applied to graphical simulation \cite{loschner2020higher}.

The rigid multibody system of robotics is a typical Hamiltonian system. 
In recent years, symplectic integrators based on Hamiltonian mechanics have attracted the attention of many researchers in robotics \cite{sharma2020review} and graphics \cite{xu2014implicit, kobilarov2009lie}. 
The semi-implicit Euler (also called symplectic Euler) method we adopt is a first-order symplectic method, which can have both performance and stability. 
Various order integrators (including explicit Euler, explicit midpoint, RK4, and RK6) are used to compare our method.
Experiments show that our method performs better in the energy conservation of long-term simulation.

\section{Methods}
\label{sec:methods}

In this section, the theoretical background and core algorithms of interactive roadheader robot simulation are described. The space algebra using 6-D vectors is introduced as the theoretical foundation of rigid body dynamics. Then, the dynamics of a single rigid body based on 6-D vector language are described. After describing the multibody dynamics, two key dynamics algorithms for our simulation system are introduced. 
Finally, the related numerical integration methods and their properties are introduced.

\subsection{Rigid Multibody Dynamics}
\label{subsec:dynamics}

\subsubsection{Spatial Algebra using 6-D Vectors}
Spatial algebra based on 6-D vectors is a concise and efficient tool for describing dynamics. In traditional computer dynamics simulation, the linear motion and rotational motion of a rigid body are represented separately. This will cause the dynamic equations to be very cumbersome and fragmented. Considering that a rigid body in three-dimensional space has six degrees of freedom (DOF), it is more appropriate to directly use six variables to represent the state of a rigid body. Some physicists and roboticists have since proposed various 6-D representation tools, such as screw theory, Lie algebra, and motor algebra.
These methods have different perspectives, but they are essentially equivalent. In this paper, we use 6-D vectors based on Plücker coordinates for easier programming. A more complete introduction is in \cite{featherstone2010beginner1}. 

The 6-D vectors does not belong to Euclidean space. There are two vector spaces as the basis of space algebra: spatial motion vector space $\bm{M}^{6}$ and spatial force vector space $\bm{F}^{6}$.
The elements of the two spaces are: spatial motion vectors $\hat{\boldsymbol{v}} \in \mathrm{M}^{6}$ and space force vectors $\hat{\boldsymbol{f}} \in \bm{F}^{6}$. The spaces $\bm{M}^{6}$ and $\bm{F}^{6}$ are equivalent to Lie algebras $\mathfrak{s e}(3)$ and $\mathfrak{s e}^{*}(3)$.
We introduce an arbitrary but fixed reference point $O$. The Plücker coordinate bases of these two vector spaces are $\mathcal{D}_{O}: \mathbb{R}^{6} \rightarrow \mathrm{M}^{6}$ and $\rm{E}_{O}: \mathbb{R}^{6} \rightarrow \rm{F}^{6}$, respectively. The $\hat{\boldsymbol{v}}$ can be written in this form:
\begin{align}
\hat{\boldsymbol{v}}=\left[\begin{array}{c}
\omega \\
v_{O}
\end{array}\right]=\left[\omega_{x}, \omega_{y}, \omega_{z}, v_{O x}, v_{O y}, v_{O z}\right]^{T},
\end{align}
where $\omega \in \mathbb{R}^{3}$ and $v_{O} \in \mathbb{R}^{3}$ represent the angular and linear velocity of a rigid body, respectively. Similarly, the $\hat{f}$ can be written in this form:
\begin{align}
\hat{\boldsymbol{f}}=\left[\begin{array}{c}
n_{O} \\
f
\end{array}\right]=\left[n_{O x}, n_{O y}, n_{O z}, f_{x}, f_{y}, f_{z}\right]^{T}
\end{align}
where $n_{O} \in \mathbb{R}^{3}$ and $f \in \mathbb{R}^{3}$ represent the total moment and linear force of a rigid body, respectively. All these variables are defined according to the reference point $O$. The essence of the spatial algebra is that a pair of $\mathbb{R}^{3}$ vectors simultaneously describe the linear and angular configuration of a rigid body.

The transformation between different coordinate frames is the key to rigid body dynamics.
$\bm{M}^{6}$ and $\bm{F}^{6}$ are dual space of each other. Correspondingly, the Plücker bases $\mathcal{D}_{O}$ and $\mathcal{E}_{O}$ define a dual coordinate system on $\bm{M}^{6}$ and $\bm{F}^{6}$. 
Let $A$ and $B$ be two Plücker coordinate system.
Let $^A \hat{\boldsymbol{f}} \in \bm{F}^{6}$, $^A \hat{\boldsymbol{m}} \in \bm{M}^{6}$ in $A$ and $^B \hat{\boldsymbol{f}} \in \bm{F}^{6}$, $^B \hat{\boldsymbol{m}} \in \bm{M}^{6}$ in $B$. The coordinate transformation for motion vectors is as follows: 
\begin{align}
{ }^{B} \hat{\boldsymbol{m}}={ }^{B} \boldsymbol{X}_{A}{ }^{A} \hat{\boldsymbol{m}}
\end{align}
where $ { }^{B} \boldsymbol{X}_{A}{ } $ is the transformation matrix from $A$ to $B$. 
Due to the duality between $\bm{M}^{6}$ and $\bm{F}^{6}$, the corresponding transformation for force vectors is called $ { }^{B} \boldsymbol{X}_{A}^{*}{ } $. The coordinate transformation for force vectors is as follows:
\begin{align}
{ }^{B} \hat{\boldsymbol{f}}={ }^{B} \boldsymbol{X}_{A}^{*}{ }^{A} \hat{\boldsymbol{f}}.
\end{align}
Let $r \in \mathbb{R}^{3}$ and $E \in \mathbb{R}^{3 \times 3}$ are position and rotation transform from $A$ to $B$, respectively. $ { }^{B} \boldsymbol{X}_{A}{ } $ and $ { }^{B} \boldsymbol{X}_{A}^{*}{ } $ are given by
\begin{align}
{ }^{B} \boldsymbol{X}_{A}{ }=\left[\begin{array}{cc}
E & 0 \\
-E r \times & E
\end{array}\right],  
\quad^{B} \boldsymbol{X}_{A}^{*}=\left[\begin{array}{cc}
E & -E r \times \\
0 & E
\end{array}\right].
\end{align}

\subsubsection{Single Rigid Body Dynamics}
A rigid body is an idealized solid model without deformation. In other words, the relative position of any pair of points of a rigid body will never change. Unlike a mass point, a rigid body has not only a linear configuration but also a rotational configuration. Therefore, at least 6 parameters are required to determine the configuration of a rigid body in $\mathbb{R}^{3}$ space. That is, the degrees of freedom of a spatial rigid body is 6. The mass of a rigid body is distributed over its volume. In physics, the inertia tensor is usually used to describe the mass distribution of rigid bodies. In the framework of spatial algebra using 6-D vectors, spatial inertia is used to replace the inertia tensor. The spatial inertia $\hat{\boldsymbol{I}}: \bm{M}^{6} \rightarrow \bm{F}^{6}$ for a rigid body is given by

\begin{align}
\hat{\boldsymbol{I}}=\left[\begin{array}{cc}
\hat{\boldsymbol{I}}_{C}+m c \times c \times^{T} & m c \times \\
m c \times^{T} & m 1
\end{array}\right]
\end{align}
where $m$ is the mass of the body, $c$ is the coordinate of the body's center of mass, and $\hat{\boldsymbol{I}}_{C}$ is the body's inertia about its center of mass. The spatial inertia is a positive-definite, symmetric $6\times6$ matrix. Analogous to transformations between different coordinate frame for motions and forces, we can define transformation for spatial inertia by:

\begin{align}
{ }^{B} \hat{\boldsymbol{I}}={ }^{B} \boldsymbol{X}_{A}^{* A} \hat{\boldsymbol{I}}^{A} \boldsymbol{X}_{B}
\end{align}
where $^{A} \hat{\boldsymbol{I}}$ in coordinate frame $A$ and $^{B} \hat{\boldsymbol{I}}$ in coordinate frame $B$. 

Given spatial inertia $\hat{\boldsymbol{I}}$ and spatial velocity $\hat{\boldsymbol{v}}$, we can define the momentum $\hat{\boldsymbol{h}} \in \bm{F}^{6}$ of a rigid body, using the formula
\begin{align}
\hat{\boldsymbol{h}}=\hat{\boldsymbol{I}} \hat{\boldsymbol{v}}.
\end{align}
Analogous to classical mechanics, acceleration is defined by the time derivative of velocity, we can define the spatial acceleration of a rigid body by the time derivative of its spatial velocity. 
Given the spatial velocity $\hat{\boldsymbol{v}}$, then the spatial acceleration is the time derivative of $\hat{\boldsymbol{v}}$:
\begin{align}
\hat{\boldsymbol{a}}=\frac{\mathrm{d}}{\mathrm{d} t} \hat{\boldsymbol{v}}=\frac{\mathrm{d}}{\mathrm{d} t}\left[\begin{array}{c}
\omega \\
v_{O}
\end{array}\right]=\left[\begin{array}{c}
\dot{{\omega}} \\
\ddot{{r}}-{\omega} \times \dot{{r}}
\end{array}\right].
\end{align}
The time derivative of the spatial inertia $\hat{\boldsymbol{I}}$ is:
\begin{align}
\frac{\mathrm{d}}{\mathrm{d} t} \hat{\boldsymbol{I}}
= \hat{\boldsymbol{v}} \times^{*} \hat{\boldsymbol{I}} - \hat{\boldsymbol{I}}\hat{\boldsymbol{v}} \times    
\end{align}
With these basic concepts of space algebra mentioned above, we can use an equation, called equation of motion, to describe the state of a rigid body in three-dimensional space. Let $\hat{\boldsymbol{v}}$, $\hat{\boldsymbol{a}}$, and $\hat{\boldsymbol{I}}$ are spatial velocity, spatial acceleration, and spatial inertia of a rigid body, respectively. The equation of motion of the rigid body is given by the formula
\begin{align} \label{EOM single}
    \hat{\boldsymbol{f}} 
    &= \frac{\mathrm{d}}{\mathrm{d} t}(\hat{\boldsymbol{h}})  \nonumber
    \\
    &= \frac{\mathrm{d}}{\mathrm{d} t}(\hat{\boldsymbol{I}} \hat{\boldsymbol{v}}) \nonumber
    \\
    &=\hat{\boldsymbol{I}} \hat{\boldsymbol{a}} + (\hat{\boldsymbol{v}} \times^{*} \hat{\boldsymbol{I}} - \hat{\boldsymbol{I}}\hat{\boldsymbol{v}} \times)\hat{\boldsymbol{v}}  \nonumber
    \\
    &=\hat{\boldsymbol{I}} \hat{\boldsymbol{a}}+\hat{\boldsymbol{v}} \times^{*} \hat{\boldsymbol{I}} \hat{\boldsymbol{v}}. 
\end{align}
where $\hat{\boldsymbol{f}}$ is the spatial force on the rigid body.

\subsubsection{Multibody Dynamics Algorithms}
A multibody system consists of rigid bodies connected by joints. In this paper, we assume that each joint connects two different rigid bodies. Each joint is essentially a holonomic kinematic constraint, which reduces the degree of freedom of the connected rigid body. A free rigid body has a degree of freedom of 6 in three-dimensional space, so it can move in any direction. The degree of freedom of a rigid body connected by a joint is reduced, which usually means that it is forbidden to move in certain directions. In fact, the effect on rigid bodies is caused by the constraint forces.


A joint is usually designed to allow 0 to 6 degrees of freedom. In our roadheader simulation system, two joints, hinge and slider, are mainly considered. The degrees of freedom of these two joints are both 1. The motion subspace matrix $\bm{S}\subset\bm{M}^{6}$ is used to define a joint constraint. Let $n_f$ be the degrees of freedom, that is, the number of constraints imposed by it is $6-n_f$, then $\bm{S}$ is $6 \times n_f$. 
The rotate (revolute) joint and slider (prismatic) joint are applied to our roadheader robot simulation system.
The detailed matrices for the joints are listed in Section \ref{sec:modeling}.


The motion of a multibody system can be described by an equation of motion like the single body above. Let $\boldsymbol{q}$, $\dot{\boldsymbol{q}}$, and $\ddot{\boldsymbol{q}}$ are generalized positions, generalized velocities, and generalized accelerations of a multibody system, the equation of motion for the system in canonical form is as follows: 
\begin{align} \label{EOM}
\boldsymbol{H}(\boldsymbol{q}) \ddot{\boldsymbol{q}}+\boldsymbol{C}(\boldsymbol{q}, \dot{\boldsymbol{q}})=\boldsymbol{\tau}
\end{align}
where $\boldsymbol{H}(\boldsymbol{q})$ is the generalized inertia matrix, $\boldsymbol{C}(\boldsymbol{q}, \dot{\boldsymbol{q}})$ is the generalized bias force (Coriolis and centrifugal forces) matrix, and $\boldsymbol{\tau}$ is the generalized external forces (e.g. gravity, user-defined joint forces, etc.). Our main goal is to solve \ref{EOM} to get the acceleration $\ddot{\boldsymbol{q}}$. We will use the Recursive Newton-Euler Algorithm (RNEA) and Composite Rigid Body Algorithm (CRBA) to get $\boldsymbol{H}$ and $\boldsymbol{C}$ respectively. Given the known variables $\boldsymbol{H}$, $\boldsymbol{C}$ and $\boldsymbol{\tau}$, we have a simple linear equation $ \boldsymbol{H}\ddot{\boldsymbol{q}} = \boldsymbol{\tau} - \boldsymbol{C}$. The acceleration $\ddot{\boldsymbol{q}}$ can be obtained directly by using any linear equation solution (such as Cholesky  decomposition  \cite{press2007numerical}). Then the numerical integrator will use $\ddot{\boldsymbol{q}}$ to get $\dot{\boldsymbol{q}}$ and $\boldsymbol{q}$ to complete one step of the simulation.

\begin{algorithm}
    \caption{Recursive Newton-Euler Algorithm}\label{RNEA}
    \begin{algorithmic}[1]

    \Require Robot state $\boldsymbol{q}, \dot{\boldsymbol{q}}$
    \Ensure Generalized bias term $\boldsymbol{C}$

    \State $\boldsymbol{v}_{0}=\mathbf{0}$
    \State $\boldsymbol{a}_{0}=-\boldsymbol{a}_{g}$

    \For{\texttt{$i=1,2,\ldots n_{B}$}}
    \State \texttt{$\left[\boldsymbol{X}_{\mathbf{J} i}, \boldsymbol{S}_{i}, \boldsymbol{c}_{\mathbf{J} i}\right]=\operatorname{jcalc}\left(\operatorname{jtype}(i), \boldsymbol{q}_{i}, \dot{\boldsymbol{q}}_{i}\right)$ }
    \State \texttt{ ${ }^{i} \boldsymbol{X}_{\lambda_{i}}=\boldsymbol{X}_{\mathbf{J} i} \boldsymbol{X}_{\mathbf{T} i}$}
    \If{ $\lambda_{i} \neq 0$ }
        \State ${ }^{i} \boldsymbol{X}_{0}={ }^{i} \boldsymbol{X}_{\lambda_{i}}{ }^{\lambda_{i}} \boldsymbol{X}_{0}$
    \EndIf
    \State $\boldsymbol{v}_{\mathbf{J} i}=\boldsymbol{S}_{i} \dot{\boldsymbol{q}}_{i}$  
    \State $\boldsymbol{a}_{\mathbf{J} i}=\boldsymbol{c}_{\mathbf{J} i}$
    \State $\boldsymbol{v}_{i}={ }^{i} \boldsymbol{X}_{\lambda_{i}} \boldsymbol{v}_{\lambda_{i}}+\boldsymbol{v}_{\mathbf{J} i}$
    \State $\boldsymbol{a}_{i}={ }^{i} \boldsymbol{X}_{\lambda_{i}} \boldsymbol{a}_{\lambda_{i}}+\boldsymbol{a}_{\mathbf{J} i}$
    \State $\boldsymbol{f}_{i}=\boldsymbol{I}_{i} \boldsymbol{a}_{i}+\boldsymbol{v}_{i} \times^{*} \boldsymbol{I}_{i} \boldsymbol{v}_{i}-{ }^{i} \boldsymbol{X}_{0}^{*} \boldsymbol{f}_{i}^{x}$
    \EndFor

    \For{\texttt{$i=n_{B},n_{B}-1,\ldots 1$}}
    \State $\boldsymbol{C}_{i}=\boldsymbol{S}_{i}^{T} \boldsymbol{f}_{i}$
    \If{$\lambda_{i} \neq 0$}
        \State $\boldsymbol{f}_{\lambda_{i}}=\boldsymbol{f}_{\lambda_{i}}+^{\lambda_{i}} \boldsymbol{X}_{i}^{*} \boldsymbol{f}_{i}$
    \EndIf
    \EndFor

    \end{algorithmic}
    \end{algorithm}

First, The Recursive Newton-Euler Algorithm (RNEA) is used to compute generalized bias term $\boldsymbol{C}$ according to the current state of the robot. The RNEA is one of the most effective inverse dynamics algorithms in robotics. It has three steps: First, compute the $\boldsymbol{q}$, $\dot{\boldsymbol{q}}$, and $\ddot{\boldsymbol{q}}$ of each rigid body. Then, compute the net force required to satisfy $\ddot{\boldsymbol{q}}$ according to \ref{EOM single} for each rigid body. Finally, compute the force transmitted across each joint. Let $\boldsymbol{a}_{g}\in \mathrm{M}^{6}$ be the acceleration of gravity in 6-D vector form. The pseudocode of Recursive Newton-Euler Algorithm is shown in Algorithm \ref{RNEA}. In this pseudocode, jtype($i$) is a function for obtaining the type of joint $i$, and jcalc() is a function for calculating joint transformation based on joint information \cite{featherstone2010beginner2}. Suppose the number of joints in the system is $n$. Let $n_B$ be the number of bodies in the system. The time complexity of RNEA algorithm is $O(n_B)$.

Then, the Composite Rigid-Body Algorithm (CRBA) is used to compute the joint generalized inertia matrix $\boldsymbol{H}$. The key process of this algorithm is to compute the non-zero part of $\boldsymbol{H}$ recursively. Specifically, the algorithm recursively calculates the matrix blocks $\boldsymbol{H}_{i j}$ and $\boldsymbol{H}_{j i}$ generated by joints $i$ and $j$. The pseudocode of CRBA is shown in Algorithm \ref{CRBA}. Refer to Section \ref{RobMod} for symbol definitions. Let $n_B$ be the number of bodies in the system. The complexity of CRBA is $O(n_B^2)$. 

Finally, we have the bias force $\boldsymbol{C}$ obtained by the RNEA, the given applied forces $\boldsymbol{\tau}$, and the $\boldsymbol{H}$ obtained by the CRBA. 
According to \ref{EOM}, we obtain a simple system of linear equations $ \boldsymbol{H}\ddot{\boldsymbol{q}} = \boldsymbol{\tau} - \boldsymbol{C}$.
Cholesky decomposition method is used to solve it directly to get the unknown acceleration $\ddot{\boldsymbol{q}}$. So far the responsibility of the dynamic algorithms has been completed. After that, the simulator will use the numerical integrator to obtain the current position $\boldsymbol{q}$ of the multibody system to complete a simulation iteration.

\begin{algorithm}
    \caption{Composite Rigid Body Algorithm}\label{CRBA}
    \begin{algorithmic}[1]

    \Require Robot physical information
    \Ensure Generalized inertia matrix $\boldsymbol{H}$
    
    \State    $\boldsymbol{H}=0$

    \For{\texttt{$i=1,2,\ldots n_{B}$}}
    \State $\boldsymbol{I}_{i}^{c}=\boldsymbol{I}_{i}$
    \EndFor

    \For{\texttt{$i=n_{B},n_{B}-1,\ldots 1$}}
    \If{$\lambda_{i} \neq 0$}
    \State $\boldsymbol{I}_{\lambda_{i}}^{c}=\boldsymbol{I}_{\lambda_{i}}^{c}+{ }^{\lambda_{i}} \boldsymbol{X}_{i}^{*} \boldsymbol{I}_{i}^{c i} \boldsymbol{X}_{\lambda_{i}}$
    \EndIf

    \State $\boldsymbol{F}=\boldsymbol{I}_{i}^{c} \boldsymbol{S}_{i}$
    \State $\boldsymbol{H}_{i i}=\boldsymbol{S}_{i}^{T} \boldsymbol{F}$
    \State $j=i$

    \While{$\lambda_{i} \neq 0$}

    \State $\boldsymbol{F}={ }^{\lambda_{i}} \boldsymbol{X}_{i}^{*} \boldsymbol{F}$ 
    \State $j=\lambda_{i}$
    \State $\boldsymbol{H}_{i j}=\boldsymbol{F}^{T}\boldsymbol{S}_{j}$
    \State $\boldsymbol{H}_{j i}=\boldsymbol{H}_{i j}^{T}$

    \EndWhile

    \EndFor

    \end{algorithmic}
    \end{algorithm}
In this subsection, we introduced the rigid multibody dynamics framework based on generalized coordinates. Based on the language of space algebra, we established the dynamics mathematical models of single rigid body and multibody. Subsequently, numerical algorithms RNEA and CRBA were used to numerically solve the dynamics mathematical model. In summary, our algorithm describes multibody dynamics with generalized coordinates, which inherently does not violate constraints. Current state-of-the-art physics engines, such as Bullet, are based on impulse-based methods \cite{coumans2013bullet}. impulse-based methods belong to the maximum coordinate method, and its basic idea is to approximate the constraints with a series of impulses \cite{bender2014interactive}. Although such methods are widely used in video games, it is difficult to meet the accuracy requirements of industrial robot simulation \cite{choi118use, liu2021role}. In recent years, Position based dynamics (PBD) method has become a new paradigm of robot simulation \cite{muller2020detailed}. However, the accuracy and stability of PBD are still inadequate. We'll show experimental comparisons in Section \ref{sec:results}.

\subsection{Numerical Integrators}
\label{subsec:integrators}
The system of ordinary differential equations constructed by rigid multibody dynamics algorithms is so complex that it has no analytical solution and must be solved iteratively by numerical integrator. The numerical integration method is always the key problem of physically-based simulation. For a given application scenario, a good numerical integrator should not only have accuracy and stability, but also have high computational performance.

The explicit Euler integrator is the most naive numerical integration method, which only utilizes the first order Taylor approximation of ordinary differential equations. Numerical instability makes explicit Euler integrator difficult to simulate complex systems. As the extensions of explicit Euler method, integrators with higher order Taylor approximations were developed, such as the explicit midpoint method and the Runge-Kutta method. In particular, the fourth-order Runge-Kutta method is widely used in the field of physical simulation. Although these methods have improved stability, as explicit methods, they still introduce numerical dissipation. We propose to use symplectic Euler integrator, which has better stability in long-term simulation.

\noindent\textbf{Explicit Euler integrator} Explicit Euler (forward Euler) integrator is the most intuitive and easy to implement method. However, the extreme instability of the explicit Euler method has been observed in the early work of graphic simulation \cite{terzopoulos1988deformable, terzopoulos1987elastically}. 
As above, we denote $\boldsymbol{q}$, $\dot{\boldsymbol{q}}$, and $\ddot{\boldsymbol{q}}$ as generalized positions, generalized velocities, and generalized accelerations, respectively.
The explicit Euler integrator has the following form:
\begin{equation}
    \left\{
    \begin{array}{lr}
    
    \dot{\boldsymbol{q}}_{n+1} = \dot{\boldsymbol{q}}_{n} + h\ddot{\boldsymbol{q}}_{n} \\
    
    \boldsymbol{q}_{n+1} = \boldsymbol{q}_{n} + h\dot{\boldsymbol{q}}_{n}
    
    \end{array}
    \right.
\end{equation}
where $h$ is the is the fixed time step length, and the subscript $n$ represents the iteration step number. Explicit Euler's method uses the first-order approximation of the current iteration to obtain the velocity and position of the next step directly. Obviously, each iteration introduces additional energy and results in a numerical explosion. Although reducing the time step $h$ can delay the occurrence of numerical explosion, the amount of calculation will be greatly increased. 
In contrast, the implicit Euler (backward Euler) method takes the first-order approximation of the next step to update the next step.That is, each iteration must solve the implicit equation.
The implicit Euler integrator is expressed as follows:
\begin{equation}
    \left\{
    \begin{array}{lr}
    
    \dot{\boldsymbol{q}}_{n+1} = \dot{\boldsymbol{q}}_{n} + h\ddot{\boldsymbol{q}}_{n+1} \\
    
    \boldsymbol{q}_{n+1} = \boldsymbol{q}_{n} + h\dot{\boldsymbol{q}}_{n+1}
    
    \end{array}.
    \right.
\end{equation}
Each iteration of the implicit Euler method introduces an energy loss, which is called artificial damping in the physics-based animation community. Although the implicit Euler method can also cause energy inaccuracy, it can guarantee that there will be no numerical explosion. Therefore, some simulation applications that require stability but do not require real time will adopt this method \cite{baraff1998large, press2007numerical}. In interactive robotics simulation, the implicit method is difficult to meet the performance requirements.

\noindent\textbf{Fourth-order Runge-Kutta method} In order to improve the accuracy of explicit Euler, various high-order explicit integration methods have been proposed. One of the most widely used high-order methods is the fourth-order Runge-Kutta method (RK4).
The time differential of generalized position and generalized velocity is expressed as $\frac{d q}{d t}=\dot{q}(t, q)$ and $\frac{d \dot{q}}{d t}=\ddot{q}(t, \dot{q})$, respectively. The form of the fourth-order Runge-Kutta integrator is as follows:
\begin{equation}
    \left\{
    \begin{array}{lr}
    
    \dot{\boldsymbol{q}}_{n+1}=\dot{\boldsymbol{q}}_{n} +
    \frac{1}{6} h\left(\ddot{k}_{1} +2 \ddot{k}_{2}+2 \ddot{k}_{3}+\ddot{k}_{4}\right) \\
    
    \boldsymbol{q}_{n+1}      =\boldsymbol{q}_{n} +
    \frac{1}{6} h\left(\dot{k}_{1} +2 \dot{k}_{2}+2 \dot{k}_{3}+\dot{k}_{4}\right) \\
    
    \end{array}.
    \right.
\end{equation}
Detailed definitions are given in Appendix \ref{appB}.

The RK4 method is much more accurate than the first order explicit Euler method. Due to RK4's efficiency and accuracy, many game engines use it for numerical integration.
However, RK4 is still an explicit integrator, so the errors will gradually accumulate in the long-term dynamics simulation. Our experimental result (Section \ref{sec:results}) shows the numerical error accumulation of RK4 method. We also choose the second-order explicit midpoint method and the sixth-order Runge-Kutta \cite{hairer1993solving} as comparison, and the numerical results are shown in Section \ref{sec:results}.

\noindent\textbf{Symplectic Euler integrator} In recent years, symplectic integration methods have attracted much attention from the physics-based simulation community. Theoretically and experimentally, symplectic integrators have good energy conservation properties for multibody systems \cite{hairer2006geometric, stern2006discrete}.
The Symplectic Euler integrator has the following form:
\begin{equation}
    \left\{
    \begin{array}{lr}
    
    \dot{\boldsymbol{q}}_{n+1} = \dot{\boldsymbol{q}}_{n} + h\ddot{\boldsymbol{q}}_{n} \\
    
    \boldsymbol{q}_{n+1} = \boldsymbol{q}_{n} + h\dot{\boldsymbol{q}}_{n+1}
    
    \end{array}.
    \right.
\end{equation}
The Symplectic Euler integrator is most notable for explicitly updating velocity but implicitly updating position. So Symplectic Euler method is also called semi-implicit Euler or semi-explicit Euler method. 
Although the second step of the  Symplectic Euler method is implicit, it is based on the result of the explicit compute of the first step. That is, Symplectic Euler method is generally first-order explicit and naturally have high computational performance.
Explicit or implicit method cause an incremental increase or loss of system energy.
But the Symplectic Euler method makes the numerical result "oscillate" around the real value \cite{hairer2006geometric, kuhl1999energy}. In this way, the energy conservation (stability) of the system can be guaranteed in the long-term simulation.
Our numerical results (Section \ref{sec:results}) show that the Symplectic Euler method is significantly more stable than other methods.


\section{Roadheader Modeling} 
\label{sec:modeling}

In this section, the method of modeling the roadheader robot is introduced. We divide the model of the roadheader into two parts: the geometric model that records static information and the robotics model that records physical information. For geometric modeling, we propose a hierarchical 3D model format. The components of the roadheader robot, the rigid bodies of components, and the geometric information of rigid bodies are recorded in this format. Correspondingly, the physical information of the robot, including the parameters of each component and joint, is modeled by the robotics model. These two models will be input into the simulation system as initial information. 


\subsection{Geometric Modeling}

\begin{figure*}[h]%
    \centering
    \includegraphics[width=0.7\textwidth]{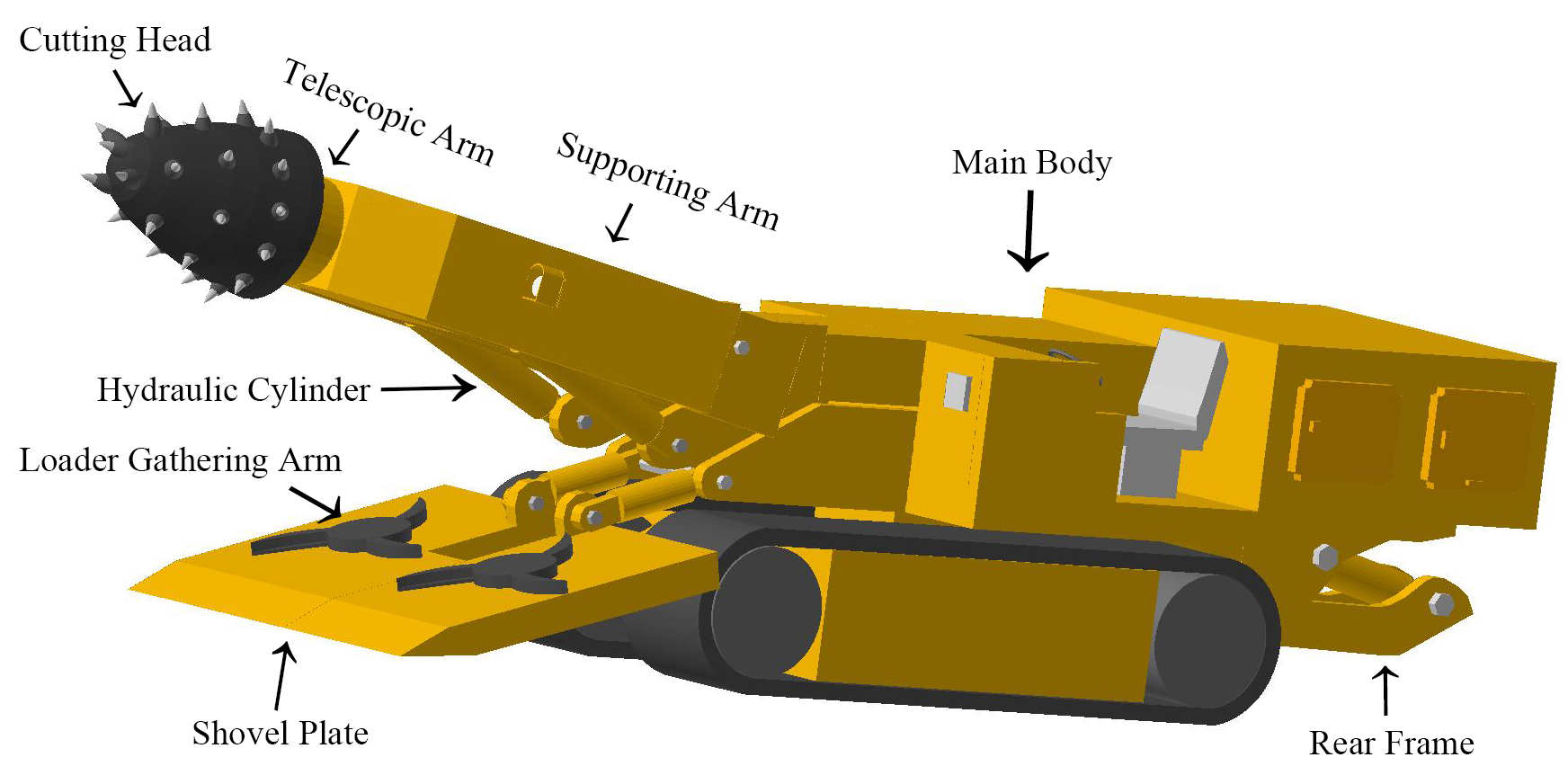}
    \caption{Roadheader Model Rendered by OpenGL.}\label{rh}
\end{figure*}

\begin{figure}[h]%
    \centering
    \includegraphics[width=0.47\textwidth]{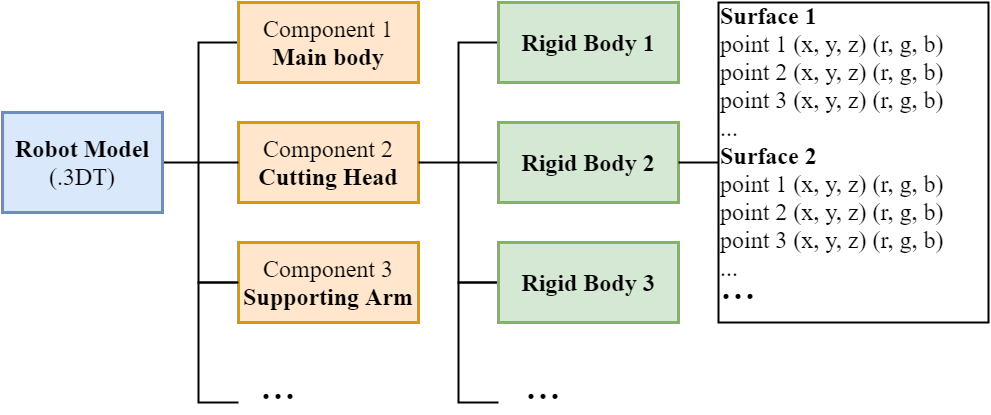}
    \caption{Hierarchical Structure of .3DT Model File.}\label{3DT}
\end{figure}

To visualize the roadheader and working environment, we propose a universal hierarchical 3D model format called .3DT. The .3DT is geometric information in plain text, which can be parsed and processed quickly and directly. Unlike the .obj format, the .3DT format can organize the robot's component models hierarchically in a single file. In our design, a roadheader robot is divided into several components like the physical world, and each component is composed of multiple rigid bodies. The .3DT file stores the hierarchical structure of a robot and the geometric information, color, and other attributes of each rigid body. The model file structure of a roadheader robot is shown in Figure \ref{3DT}.

In this paper, we will use an EBZ230 roadheader as a prototype to implement the simulation system. The .3DT format 3D model of this roadheader is rendered with OpenGL as shown in Figure \ref{rh}. The entire robot model consists of 162 rigid body models with a total of 24582 points and 27233 surface elements. The key components of the roadheader model are also marked in Figure \ref{rh}. This research only focuses on the movement of important functional components (such as cutting parts and shovel) and ignore other details (such as crawlers) of the robot. 
\begin{figure}[h]%
    \centering
    \includegraphics[width=0.35\textwidth]{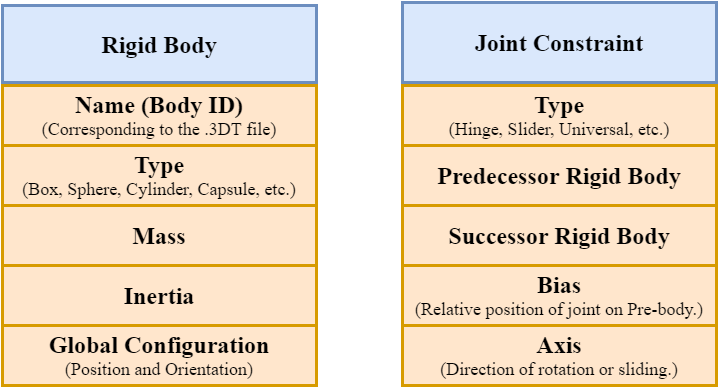}
    \caption{The data structure of Components and Joints.}\label{RobMod}
\end{figure}

\subsection{Robotics Modeling}

The geometric modeling of the roadheader robot only has static graphics information. To describe the physical information of the robot, the robotics modeling is introduced into our system. Robotics modeling consists of two important parts: the physical information of each robot component and the constraints between the two components. The former includes physical information such as the mass and inertia of each component, which corresponds to the components in the geometric model. The latter includes the constraint information between two robot components, that is, the description of a joint.


The description of the data structure of the robot components and joints is shown in Figure \ref{RobMod}. A rigid body is the basic element in a robot system, and its data structure consists of Name (the ID corresponding to the .3DT file), Type, Mass and Inertia, Global Configuration (global position and orientation), etc. In our design, complex components (such as the Cutting Head) of the robot are composed of various rigid bodies connected by fixed joints. Therefore, the entire simulation system only contains a few basic rigid body types, such as Box, Sphere, Cylinder, Capsule, etc. The data structure of the joint constraint is consists of Type (including Hinge, Slider, Universal, Fixed, etc.), Predecessor and Successor Rigid Body, Bias (the relative position of the joint on Predecessor Rigid Body), Axis (the rotation axis or sliding direction of the joint), etc. Both the robotics model file and the geometric model file will be parsed by the simulator as the initial data of the dynamics algorithm.

\begin{figure}[h]%
    \centering
    \includegraphics[width=0.47\textwidth]{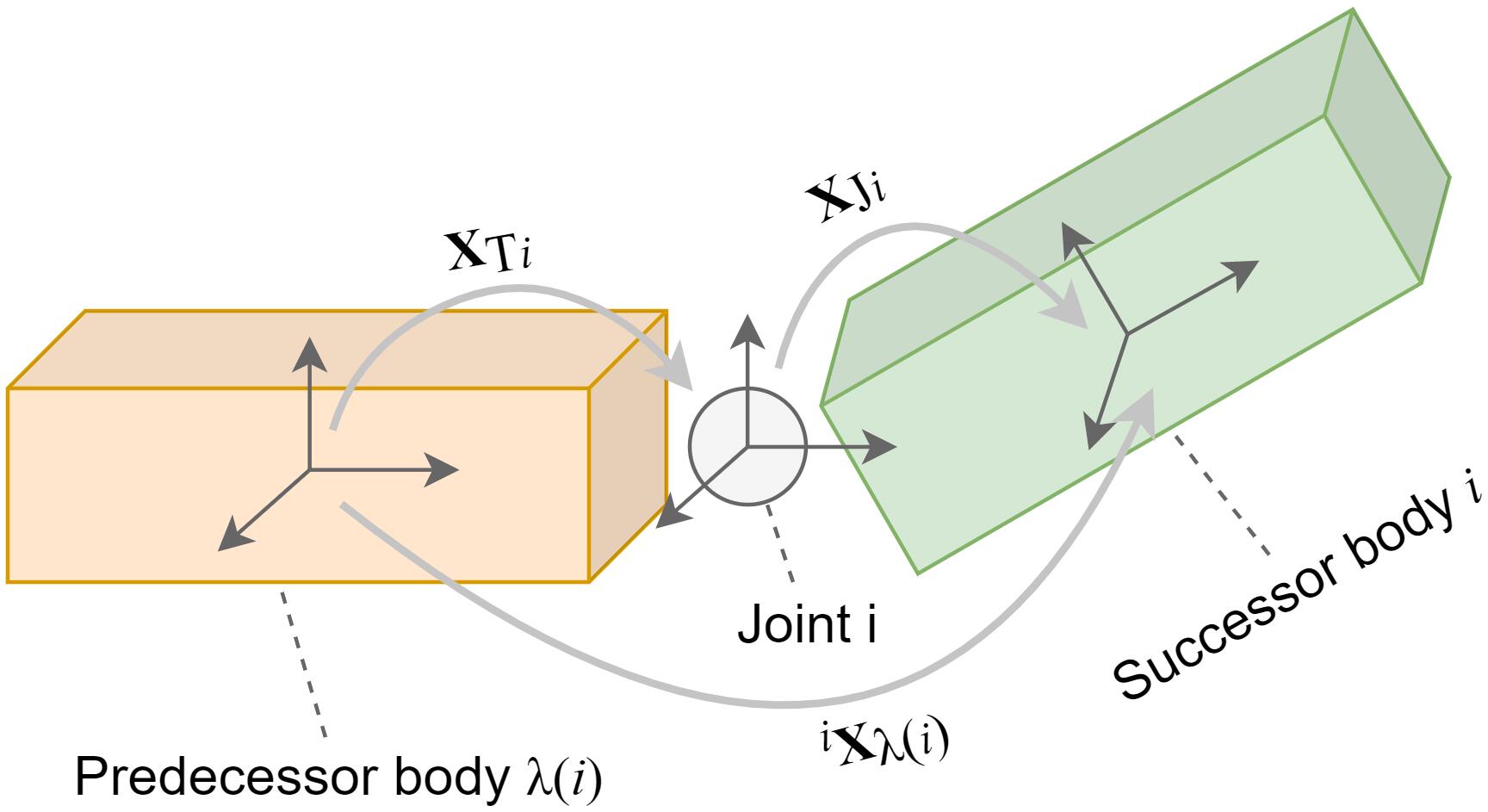}
    \caption{The predecessor and successor rigid body are connected by a joint. The motion space matrix for joint $i$ is $\boldsymbol{S}_{i}$. The spatial transformation matrices of the bodies and joint are shown in the figure. The spatial force $\boldsymbol{f}_{i}$ from body $\lambda_{i}$ is applied to body $i$ via joint $i$.}\label{joint}
\end{figure}

Each joint connects two rigid bodies: the predecessor and successor. The relationship between joint and rigid bodies is shown in Figure \ref{joint}. The predecessor and successor of any joint $i$ are called body $\lambda_{i}$ and body $i$, respectively. The spatial transformations from body $\lambda_{i}$ to joint $i$ and from joint $i$ to body $i$ are denoted as $\boldsymbol{X}_{\mathbf{T} i}$ and $\boldsymbol{X}_{\mathbf{J} i}$, respectively. The spatial velocity, spatial acceleration, spatial inertia and composite inertia of body $i$ are $\boldsymbol{v}_{i}$, $\boldsymbol{a}_{i}$, $\boldsymbol{I}_{i}$ and $\boldsymbol{I}_{i}^{c}$ respectively. The corresponding notations for body $\lambda_{i}$ is similar. The spatial force $\boldsymbol{f}_{i}$ from body $\lambda_{i}$ is applied to body $i$ via joint $i$. The motion subspace matrix $\boldsymbol{S}_{i}$ is used to describe the constraint action of joint $i$ on the configuration space of the rigid bodies. For the roadheader robot simulation, we use two kinds of joints: rotate (revolute) joint and slider (prismatic) joint. 
The motion subspace matrix of the rotate joint with the x-axis, y-axis and z-axis as the rotation axis are defined as
$$
\boldsymbol{S}_{R x}=\left[\begin{array}{l}
1 \\0 \\0 \\0 \\0 \\0
\end{array}\right], \boldsymbol{S}_{R y}=\left[\begin{array}{l}
0 \\1 \\0 \\0 \\0 \\0
\end{array}\right], \boldsymbol{S}_{R z}=\left[\begin{array}{c}
0 \\0 \\1 \\0 \\0 \\0
\end{array}\right] ,
$$
respectively. The motion subspace matrix of slider joint with the x-axis, y-axis and z-axis as the motion direction are defined as
$$
\boldsymbol{S}_{T x}=\left[\begin{array}{l}
0 \\0 \\0 \\1 \\0 \\0
\end{array}\right], \boldsymbol{S}_{T y}=\left[\begin{array}{l}
0 \\0 \\0 \\0 \\1 \\0
\end{array}\right], \boldsymbol{S}_{T z}=\left[\begin{array}{l}
0 \\0 \\0 \\0 \\0 \\1
\end{array}\right] ,
$$
respectively.

\begin{figure}[h]%
    \centering
    \includegraphics[width=0.47\textwidth]{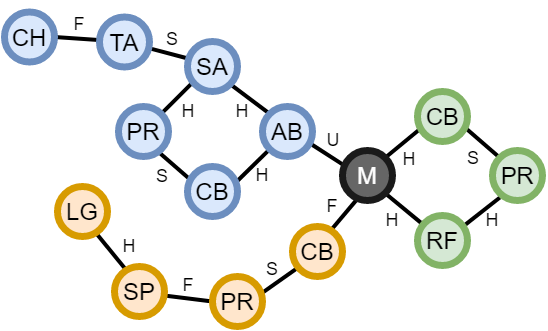}
    \caption{Topology of the Roadheader Robot. Description of nodes: M (Main Body) CH (Cutting Head), TA (Telescopic Arm), SA (Supporting Arm), AB (Arm Base), PR (Piston Rod), CB (Cylinder Barrel), RF (Rear Frame), SP (Shovel Plate), LG (Loader Gathering Arm). Description of arcs: H (Hinge Joint), S (Slider Joint), F (Fixed Joint), U (Universal Joint, equivalent to the cascade of two hinge joints).}\label{Topology}
\end{figure}

All joints and components information constitute the topology of the entire Roadheader robot. The topology graph of the key structure the Roadheader is shown in Figure \ref{Topology}. In this graph, nodes represent robot components, and arcs represent joints. The simulation system will compute the real-time state of each joint and component frame by frame according to the dynamic algorithm. Just like manipulating a robot in reality, the user sends instructions (via control panel) to change the state of the joints so that the corresponding components move.


\section{System Architecture and Implementation}
\label{sec:implement}

\begin{figure}[h]%
    \centering
    \includegraphics[width=0.47\textwidth]{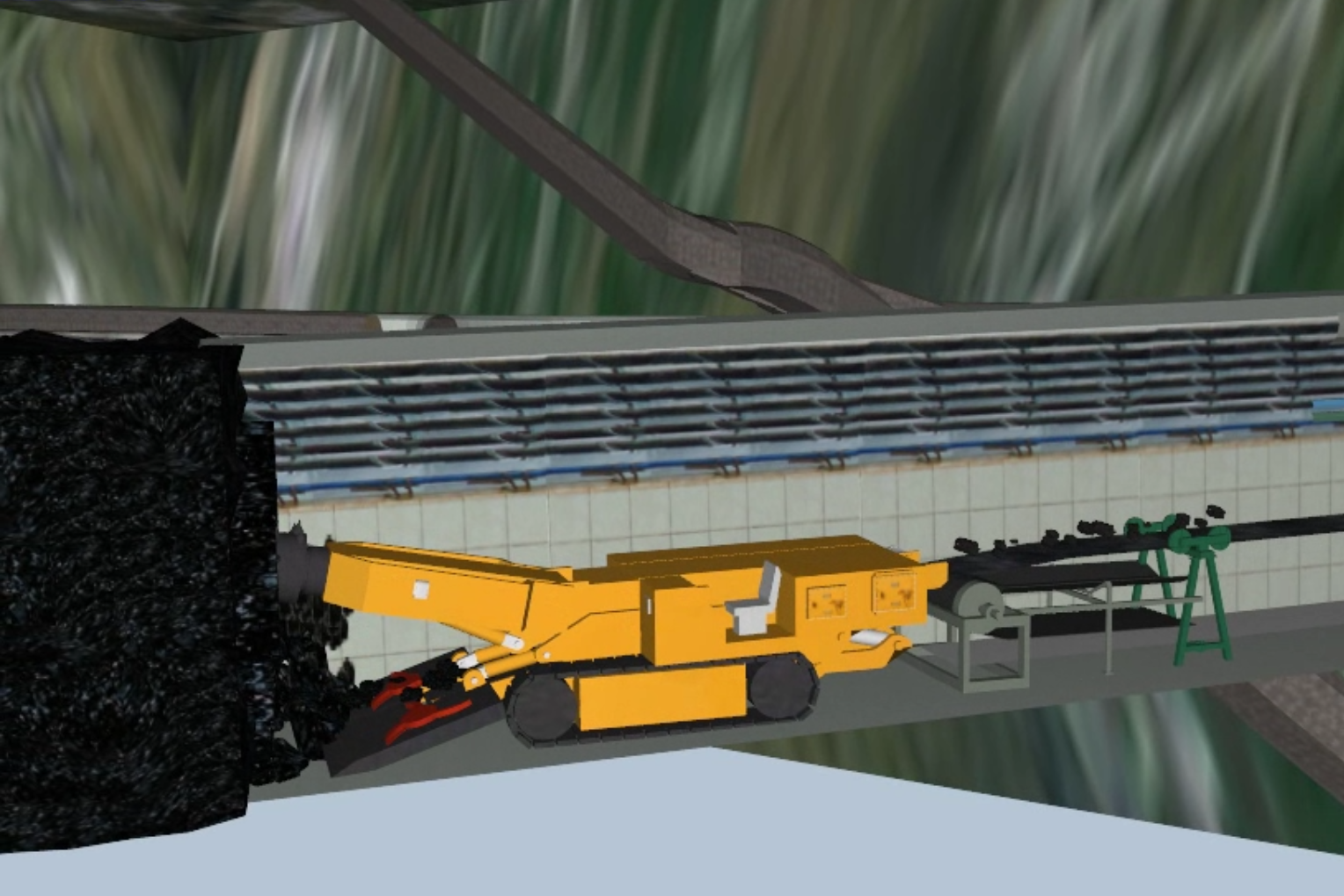}
    \caption{Roadheader Simulation Visualization Effect.}\label{Environment}
\end{figure}

\begin{figure}[h]%
    \centering
    \includegraphics[width=0.47\textwidth]{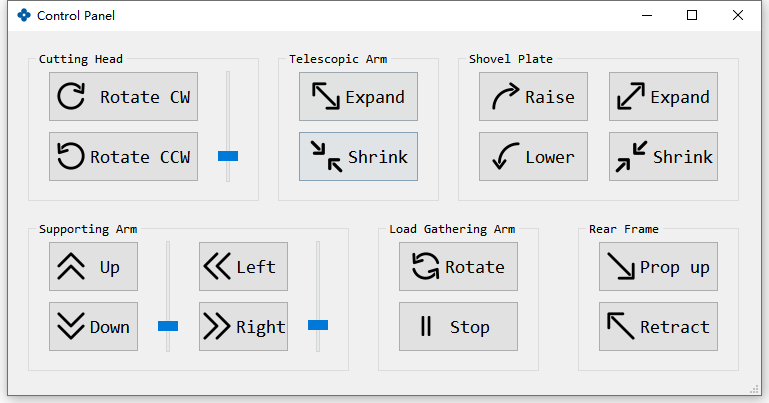}
    \caption{Control Panel.}\label{ControlPannel}
\end{figure}

The architecture of our roadheader simulation system is shown in Figure \ref{framework}. The whole system consists of two main parts: the Graphical User Interfaces (GUI) on the front end and the simulator on the back end. The GUI includes a renderer (implemented by OpenGL) to display the computation results from the simulator in real time. The rendering result of a roadheader working in an underground space is shown in Figure \ref{Environment}. Another major part of the GUI is the control panel, which accepts user instructions (via keyboard and mouse). The control panel will send user control signals (such as raising the supporting arm) to the simulator, or sending rendering commands (such as changing the viewport) directly to the renderer. A demo of the control panel is shown in Figure \ref{ControlPannel}.



\begin{algorithm}
    \caption{Interactive Roadheader Simulation Loop}\label{algo}
    \begin{algorithmic}[1]

    \Require Geometric model, Robotics model
    \Ensure Simulation visualization results

    \State $h \leftarrow \Delta t / numSubsteps $;
    \State  $\boldsymbol{q} \leftarrow initial position, \dot{\boldsymbol{q}}  \leftarrow \boldsymbol{0}, \ddot{\boldsymbol{q}} \leftarrow \boldsymbol{0} $;
    \State $currentState  \leftarrow (\boldsymbol{q}, \dot{\boldsymbol{q}}, \ddot{\boldsymbol{q}})$;

    \While{simulating}
    \If{userAction == quitSim }
    \State Quit the simulation loop;
    \EndIf

    \For{$numSubsteps$}
    \If{$userAction \neq null$}
        \State $\ddot{\boldsymbol{q}}  \leftarrow$ 
        \State $Solver(currentState, userAction)$;
    \Else
        \State $\ddot{\boldsymbol{q}}  \leftarrow Solver(currentState)$;
    \EndIf


    \State $\dot{\boldsymbol{q}} \leftarrow \dot{\boldsymbol{q}}+h \ddot{\boldsymbol{q}}$;
    \State $\boldsymbol{q} \leftarrow \boldsymbol{q}+h \dot{\boldsymbol{q}}$;
    \State $currentState  \leftarrow (\boldsymbol{q}, \dot{\boldsymbol{q}}, \ddot{\boldsymbol{q}})$;

    \EndFor
    \State Renderer($\boldsymbol{q}$);

    \EndWhile

    \end{algorithmic}
    \end{algorithm}


The process of the interactive simulation loop is shown in Algorithm \ref{algo}.
The model files that are input into the simulator consists of the two parts, geometric model and robotics model, introduced in Section \ref{sec:modeling}.
After initializing the current state (step 2 and 3), the main simulation loop begins.
Each time step $\Delta t$ is divided into several substeps $h$.
In each substep, the simulator will compute (denoted as $Solver$ in step 11 and 13) the current acceleration $\ddot{\boldsymbol{q}}$ based on the last state and the user's command using dynamics algorithms (introduced in Section \ref{sec:methods}). Then the numerical integrator will compute the current velocity $\dot{\boldsymbol{q}}$ and position $\boldsymbol{q}$ based on the current acceleration (step 15, 16, and 17). 
The simulation state will be rendered at each time step.

\section{Numerical Results}
\label{sec:results}

In this section, we designed a series of experiments to test the numerical results of our roadheader simulation system. We tested three main areas: performance, accuracy, and energy conservation. A personal computer with a Core-i5 CPU at 3.4 GHz and a GeForce GTX 950M GPU was used to test our roadheader simulation system. The whole system is implemented in C++ with Visual Studio 2019 and runs on Windows 10. The rendering effect of the system is based on OpenGL and is embedded in a Qt5 window. In all tests, the acceleration of gravity was set to $9.8 m / s^{2}$. 

\begin{figure}[h]%
    \centering
    \includegraphics[width=0.47\textwidth]{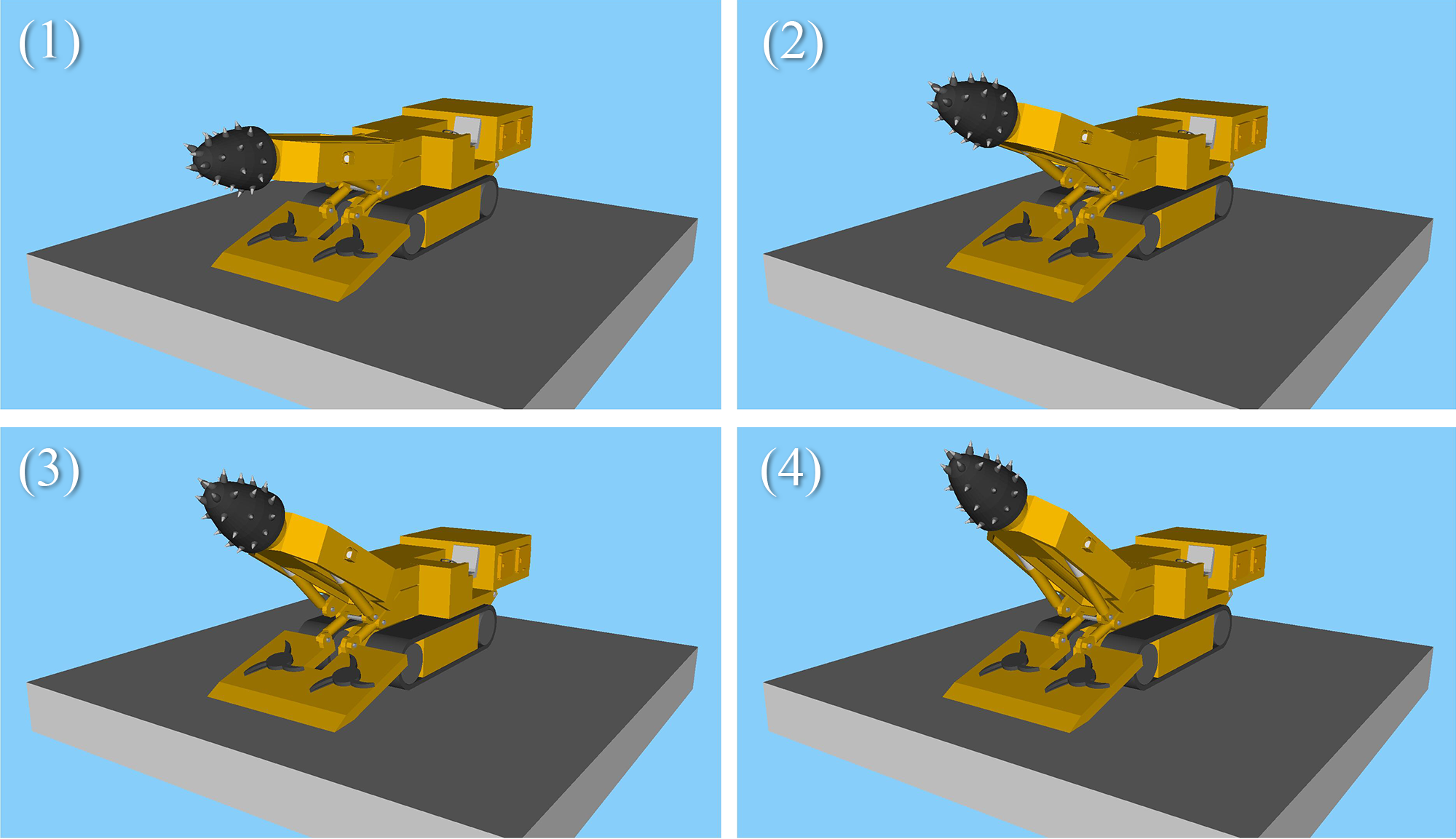}
    \caption{The Process of Raising the Supporting Arm of Roadheader.}\label{rising}
\end{figure}

\subsection{Accuracy}
Compared with video games, industry-oriented simulation systems require higher simulation accuracy. The position error (or called position drift) during the simulation is used to measure the accuracy of the simulator. However, complex robot models usually have no analytical solution, so the specific position error of the simulator is difficult to obtain. Therefore, we test the position drift through a simple arm raising motion. The process of raising the supporting arm of roadheader is shown in Figure \ref{rising}. This movement process is a simple circular movement with a closed-form solution,  that is, the distance between the cutting head and the main body is constant. Let $d$ be the initial distance, and $d'$ be the current distance computed frame by frame. Then we get the percentage of position error for each iteration:
$$ \frac{\|d-d'\|}{d'} \times 100 \% $$

\begin{figure}[h]%
    \centering
    \includegraphics[width=0.47\textwidth]{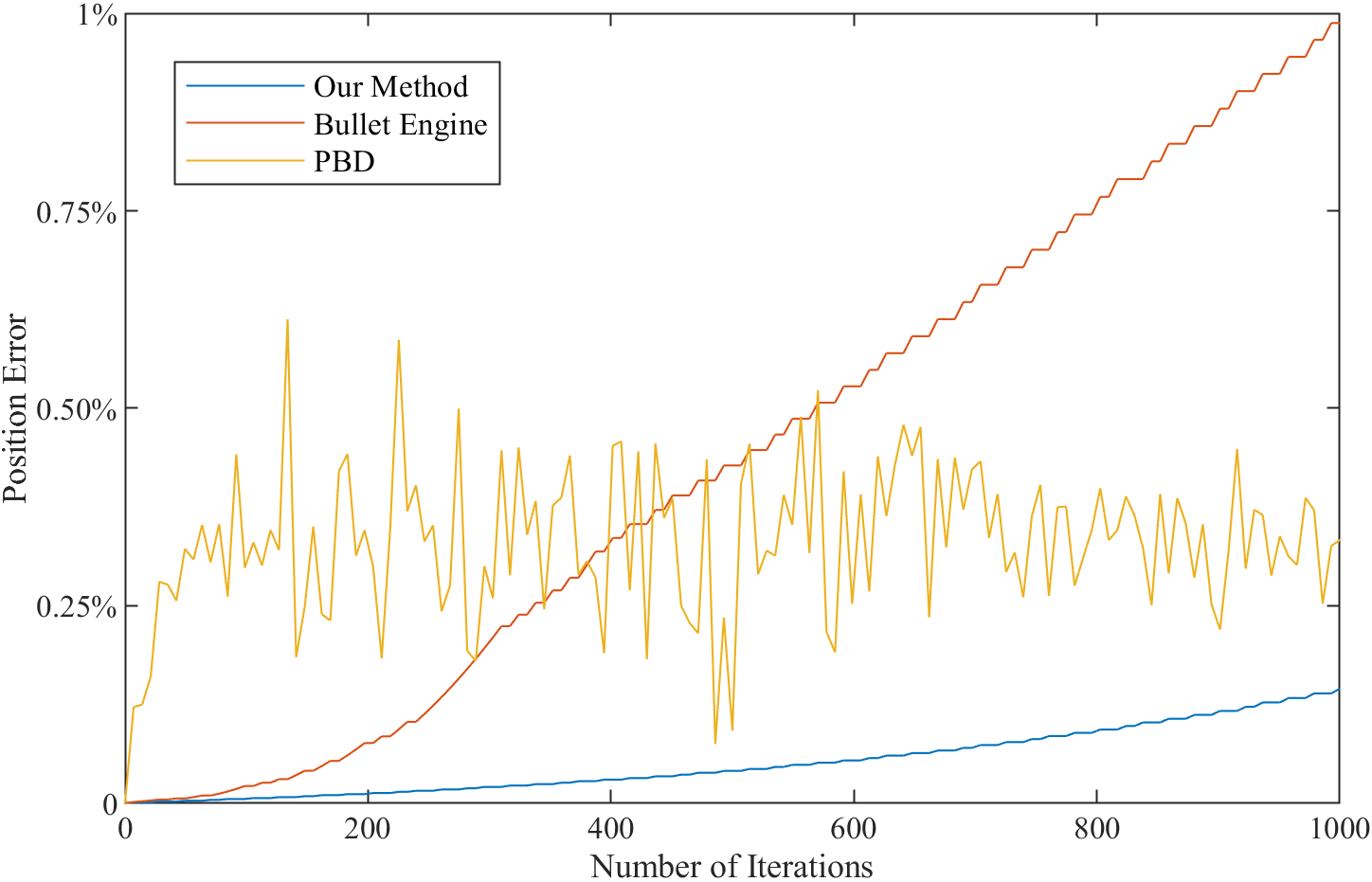}
    \caption{Relationship between iterations number and position error in different simulators.}\label{position}
\end{figure}

We use the same robot model to run in our simulation system (using generalized coordinate algorithms), Bullet physics engine (using maximal coordinate algorithms) and Position based dynamics (PBD) method \cite{muller2020detailed} to compare position errors. The contrast results of position error and the number of iterations between our method and Bullet engine is shown in Figure \ref{position}. The Bullet engine using the maximal coordinate algorithm has a significantly faster position error growth than our simulator using the generalized coordinate algorithm. The PBD method has significant fluctuation of position error, which means its stability and accuracy are not reliable. It turns out that our method has more advantages in terms of accuracy.



\subsection{Energy conservation} We explored the influence of different numerical integrators on the simulation results. In our simulation system, symplectic Euler method is used instead of conventional Runge–Kutta method. In most graphics applications, the higher the order of the integrator, the higher the accuracy of the simulation (and the higher the computational cost). However, this rule is not necessarily correct for industrial-oriented machine simulation. In theory, the total mechanical energy (the sum of the potential and kinetic energies) of a multibody system that is only acted on by conservative forces remains constant. But in the process of numerical simulation, every iteration will inevitably lead to truncation errors, that is, cause energy drift of the simulation system. Therefore, energy conservation is very important for long-term simulation. Although the RK4 method has higher accuracy in short-term simulation, it will cause energy drift accumulation like explicit Euler method. The symplectic Euler method used in our system is an energy preserving method, which performs well in long-term simulations.

\begin{figure}[ht]%
    \centering
    \includegraphics[width=0.47\textwidth]{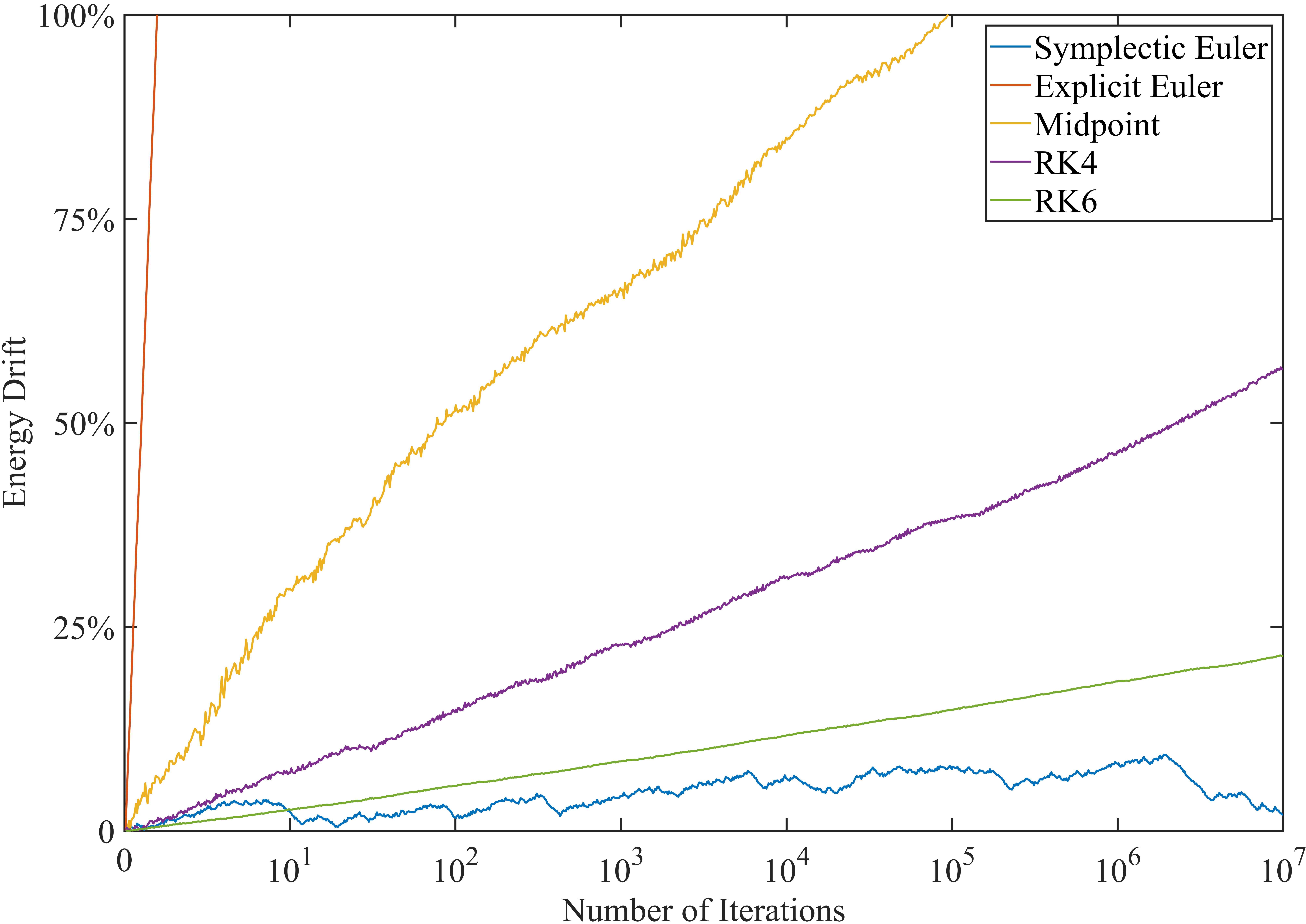}
    \caption{Relationship between iterations number and energy drift under different time integrators.}\label{integrator}
\end{figure}

We designed an experiment to compare the energy drift of symplectic Euler method, explicit Euler method (first order), explicit midpoint method (second order), RK4 method (fourth-order,), and RK6 method (sixth order) in the same environment.
Let $E$ and $E'$ be the current and initial total energy (the Hamiltonian), respectively. The percentage of energy drift 
$$\frac{\|E-E'\|}{E'} \times 100 \% $$
is computed frame by frame. 
The relationship between the number of iterations and energy drift under different time integrators is shown in Figure \ref{integrator}.
As shown in Figure \ref{integrator}, low order methods (such as explicit Euler and explicit midpoint) are easy to make the system energy drift significantly. So the lower order methods is obviously unstable and cannot be used for long-term simulation. 
The medium order method (RK4) performs well in short time but has poor energy conservation in long-term simulation.
The higher order method (RK6) performs better, but higher order means more computation. Our experiments show that its efficiency is too low to meet the needs of interactivity. 
As shown in Figure \ref{integrator}, in the long-term simulation, the energy drift of the symplectic Euler will stabilize within a small range.

Experiments show that the symplectic Euler method adopted by our system is more accurate and stable for multibody simulation than other non-preserving numerical integrators.


\subsection{Performance}
\begin{figure}[h]%
    \centering
    \includegraphics[width=0.47\textwidth]{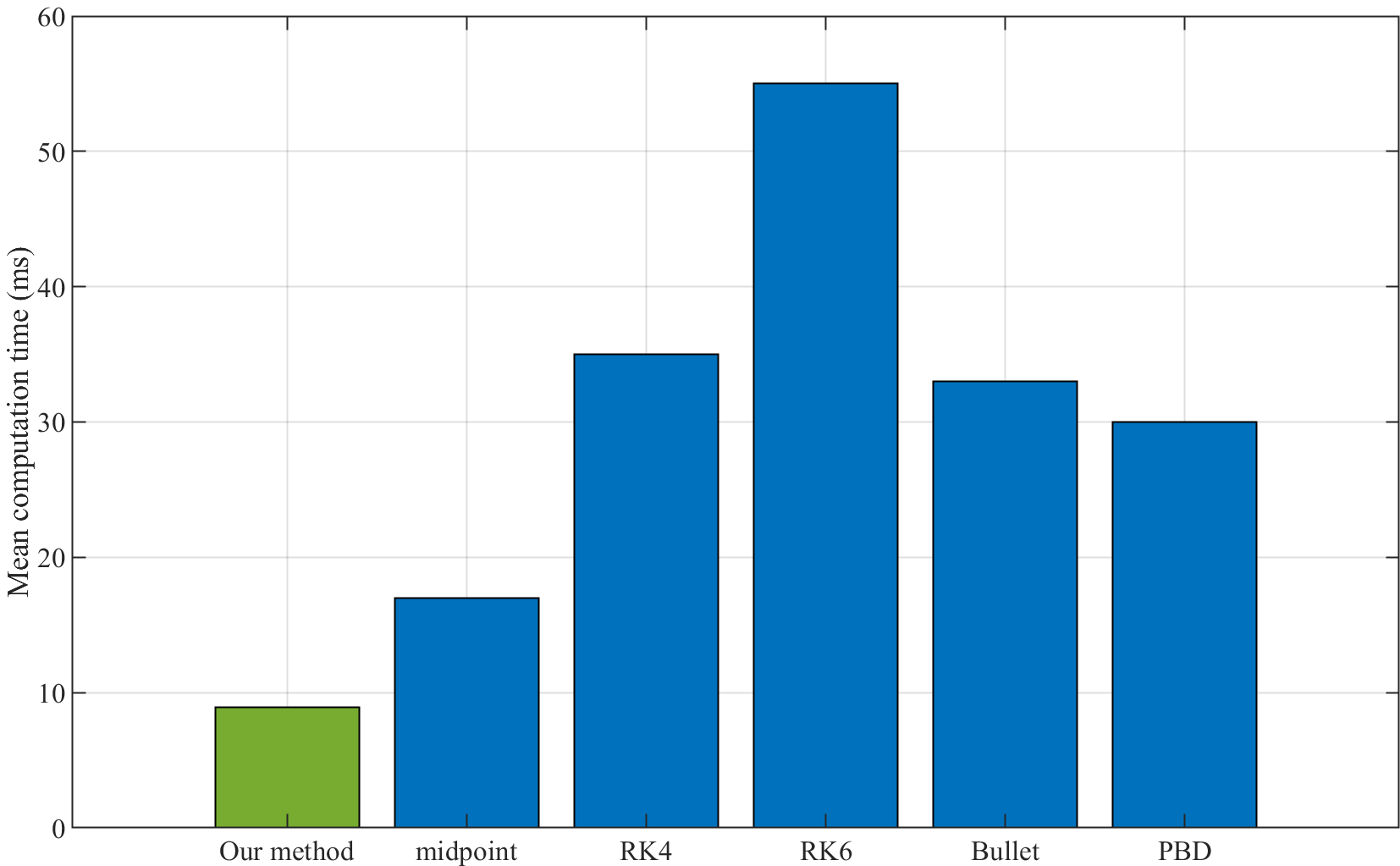}
    \caption{Comparison of computational performance of different methods.}\label{bar}
\end{figure}

We tested the performance of our simulation system. A model of an EBZ230 roadheader with 4582 points and 27233 surface elements was used in our test system. The rendering of the system is based on OpenGL 4.5 with a resolution of $1920 \times 1080$. The simulator should be fast enough to support real-time interaction between the user and the simulation system. Usually, a user-friendly simulator requires the system to reach at least 60 fps, which means that each frame takes at most 16.67 milliseconds (ms). We use the mean computation time per frame as the performance metric. The test results of our method and others are shown in Figure \ref{bar}. According to multiple tests, the simulation system takes about 8.9 ms per frame, which meets the need for Real-time interaction. Compared with other methods, our method has better computational performance.

\section{Limitations and Future Work}
\label{sec:lim}

Compared with other types of methods (maximal coordinates methods, PBD), although the numerical error of our method is smaller, its mathematical theory is more difficult to understand and code. In future work, we will try to use more intuitive mathematical description forms. The integrator we use is more stable than the other non-preserving numerical integrators , but there is still the inevitable energy drift phenomenon. In the future we will investigate more advanced forms of integration (such as implicit midpoint and Newmark integrator) or use variable time steps to further reduce energy drift.

In this paper, only multibody dynamics algorithms are used to model the roadheader robot. The scenario in which the robot works is much more complex in reality. In future work, we will study the interaction between the roadheader robot and the environment, such as impact, contact and friction. And we will improve the usability of the system from the perspective of human-computer interaction. Our method can also be extended to other types of robots. We hope our work will inspire further work that combines graphics and industry.

\section{Conclusion}
\label{sec:conclusion}
This paper presents an accurate and stable interactive simulation system for roadheader robot.
The roadheader robot is modeled in two parts: the robotics model for physical information and the geometric model for visual information. The simulator constructs the equation of motion of roadheader model based on generalized coordinates. In each iteration, the simulator computes the new state of the robot based on previous state and user interaction, and renders it in the user interface.

 
First, existing works in the field of roadheader simulation are based on commercial game engines, and the accuracy cannot meet industrial requirements. We adopted the generalized coordinates based algorithms to simulate the roadheader robot, which is more accurate than other state-of-the-art methods. We compared the proposed system to the conventional Bullet game engine. The simulation results show that the position error of our method is lower for the same simulation process. Second, the time integrator has a significant effect on simulation stability. We adopted the symplectic Euler integrator instead of the most common RK4 method. Our approach is a better trade-off between efficiency and energy accuracy. Common integrators of different orders are used for comparison. Numerical tests show that our method has less energy drift in long-term simulations. That is, our method is more stable.
In the end, our system reached 60 fps, which meets the requirements of an interactive simulation system.

\bmhead{Acknowledgments}
This work was supported in part by the National Key Research and Development Projects of China under Grant 2017YFC0804406, and in part by the National Natural Science Foundation of China under Grant 51904173.

\begin{appendices}

\section{Symbols} \label{appA}

The symbol $\times$ and $\times^{*}$ are spatial cross product operators. For any vector $v = \left[x, y, z\right]^\mathsf{T}$:
\begin{align*}
v \times=\left[\begin{array}{c}
x \\
y \\
z
\end{array}\right] \times=\left[\begin{array}{ccc}
0 & -z & y \\
z & 0 & -x \\
-y & x & 0
\end{array}\right], 
\end{align*}
\begin{align*}
v \times^{*} = - v \times^\mathsf{T}.
\end{align*}

Other important symbols and descriptions are listed in Table \ref{symbols}.

\begin{table}[htbp] \label{symbols}
    \caption{List of Symbols}
    \begin{tabular}{@{}ll@{}}
    \toprule
    Symbol & Description  \\ \midrule
    $\hat{\boldsymbol{v}}_{i}$ & Spatial velocity of body $i$ \\
    $\hat{\boldsymbol{a}}_{i}$ & Spatial acceleration of body $i$ \\
    $\hat{\boldsymbol{f}}_{i}$ & Spatial force acting on body $i$ \\
    $\hat{\boldsymbol{I}}_{i}$ & Spatial inertia of body $i$ \\
    $\hat{\boldsymbol{h}}_{i}$ & Spatial momentum of body $i$ \\
    ${}^{j} \boldsymbol{X}_{i}$ & Spatial transformation from frame $i$ to frame $j$ \\ 
    $\lambda_{i}$ & Predecessor of joint $i$ \\
    $\boldsymbol{q}, \dot{\boldsymbol{q}}, \ddot{\boldsymbol{q}}$ & Generalized position, velocity and acceleration \\ 
    $\boldsymbol{H}$ & Generalized inertia \\
    $\boldsymbol{C}$ & Generalized bias force \\
    $\boldsymbol{\tau}$ & Generalized external force \\
    $\boldsymbol{S}_{i}$ & Motion subspace matrix \\
    $h$ & Time step of integration \\ 
    \bottomrule
    \end{tabular}
\end{table}

\section{Details of RK4 method}\label{appB}
The form of the fourth-order Runge-Kutta integrator is as follows:
\begin{equation*}
    \left\{
    \begin{array}{lr}
    
    \dot{\boldsymbol{q}}_{n+1}=\dot{\boldsymbol{q}}_{n} +
    \frac{1}{6} h\left(\ddot{k}_{1} +2 \ddot{k}_{2}+2 \ddot{k}_{3}+\ddot{k}_{4}\right) \\
    
    \boldsymbol{q}_{n+1}      =\boldsymbol{q}_{n} +
    \frac{1}{6} h\left(\dot{k}_{1} +2 \dot{k}_{2}+2 \dot{k}_{3}+\dot{k}_{4}\right) \\
    
    \end{array}.
    \right.
\end{equation*}
where 
\begin{equation*}
    \left\{
    \begin{array}{lr}
        \ddot{k}_{1}=\ddot{\boldsymbol{q}}\left(t_{n}, \dot{\boldsymbol{q}}_{n}\right) \\
        \ddot{k}_{2}=\ddot{\boldsymbol{q}}\left(t_{n}+\frac{h}{2}, \dot{\boldsymbol{q}}_{n}+h \frac{\ddot{k}_{1}}{2}\right) \\ 
        \ddot{k}_{3}=\ddot{\boldsymbol{q}}\left(t_{n}+\frac{h}{2}, \dot{\boldsymbol{q}}_{n}+h \frac{\ddot{k}_{2}}{2}\right) \\
        \ddot{k}_{4}=\ddot{\boldsymbol{q}}\left(t_{n}+h, \dot{\boldsymbol{q}}_{n}+h \ddot{k}_{3}\right) \\ 
    \end{array}
    \right.
\end{equation*}
and
\begin{equation*}
    \left\{
    \begin{array}{lr}
        \dot{k}_{1}=\dot{\boldsymbol{q}}\left(t_{n}, {\boldsymbol{q}}_{n}\right) \\
        \dot{k}_{2}=\dot{\boldsymbol{q}}\left(t_{n}+\frac{h}{2}, {\boldsymbol{q}}_{n}+h \frac{\dot{k}_{1}}{2}\right) \\
        \dot{k}_{3}=\dot{\boldsymbol{q}}\left(t_{n}+\frac{h}{2}, {\boldsymbol{q}}_{n}+h \frac{\dot{k}_{2}}{2}\right) \\
        \dot{k}_{4}=\dot{\boldsymbol{q}}\left(t_{n}+h, {\boldsymbol{q}}_{n}+h \dot{k}_{3}\right) \\
    \end{array}.
    \right.
\end{equation*}

\end{appendices}



\bibliography{references}


\end{document}